\pdfoutput=1
\PassOptionsToPackage{table}{xcolor}
\documentclass{article} 
\usepackage{iclr2027_conference,times}

\usepackage{amsmath,amsfonts,bm}

\def\eqref#1{equation~\ref{#1}}

\def\1{\bm{1}}

\DeclareMathAlphabet{\mathsfit}{\encodingdefault}{\sfdefault}{m}{sl}
\SetMathAlphabet{\mathsfit}{bold}{\encodingdefault}{\sfdefault}{bx}{n}

\usepackage{hyperref}
\usepackage{url}
\usepackage{booktabs}
\usepackage{graphicx}
\usepackage{caption}
\usepackage{booktabs}
\usepackage{multirow}
\usepackage{xcolor}
\usepackage{enumitem}
\usepackage[export]{adjustbox} 

\definecolor{bestcyan}{HTML}{C1E1C1}
\definecolor{secondcyan}{RGB}{211,246,248}

\usepackage{graphicx}
\usepackage{wrapfig}
\usepackage{tabularx}
\usepackage{amsmath}
\usepackage{amssymb}
\usepackage{bm}
\usepackage{natbib}
\usepackage{algorithm}
\usepackage{algorithmic}

\title{PhyCAD: Physics-Aware Text-to-CAD Generation via Hierarchical Reinforcement Learning}

\author{
\textbf{Yuxin Cui},\,
\textbf{Weijie Zheng},\,
\textbf{Zhengyang Li},\,
\textbf{Han Wang},\,
\textbf{Wenjie Feng},\,
\textbf{Qiyu Kang}\thanks{Corresponding author: \texttt{qiyukang@ustc.edu.cn}} \\[0.5em]
University of Science and Technology of China, Hefei, China \\[0.3em]
}

\iclrfinalcopy 
\begin{document}

\maketitle
\fancyhead{}
\fancyhead[L]{Preprint}

\begin{figure}[h]
    \centering
    \includegraphics[width=\textwidth]{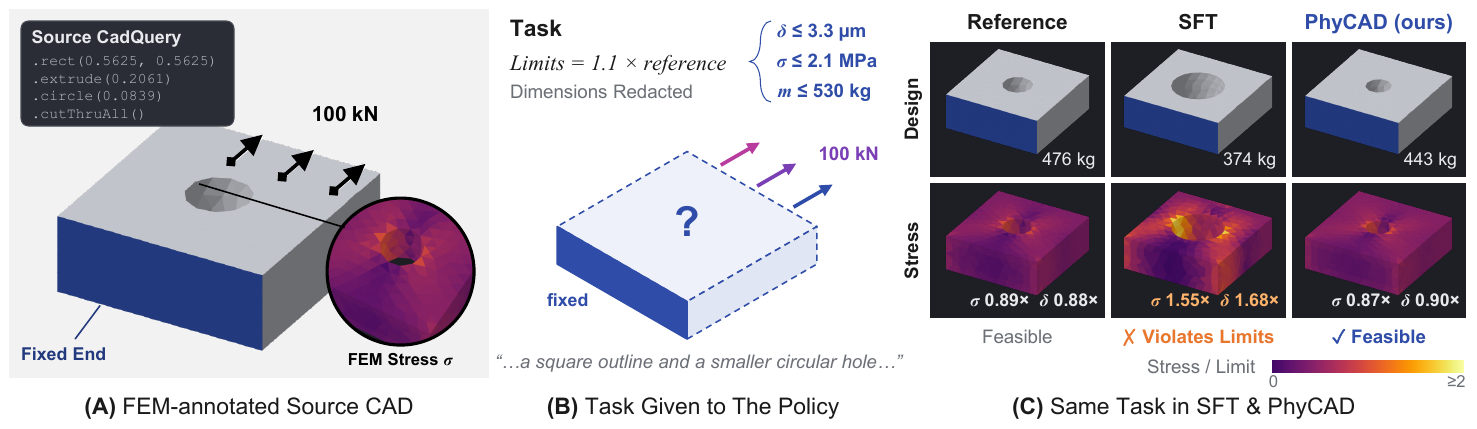}
    \caption{
\textbf{Physics-aware CAD optimization.}
\textbf{(A)} A source CAD model is annotated with FEM conditions and responses.
\textbf{(B)} Its exact dimensions are redacted to form an inverse-design task with physical constraints.
\textbf{(C)} While the SFT design achieves lower mass but violates the physical constraints, PhyCAD produces a feasible design with lower mass than the reference.
}
    \label{fig:teaser}
\end{figure}

\begin{abstract}
Language-driven computer-aided design (CAD) generation enables the creation of editable parametric models from text, yet geometric correctness alone does not ensure physical feasibility.
We present PhyCAD, a framework for physics-aware Text-to-CAD generation that trains language models to produce executable CadQuery programs under geometric and physical constraints.
To support this setting, we construct a Text-to-CAD dataset that links parametric CAD models with geometric specifications, loading conditions, and physical constraints.
We further develop an automated CAD--CAE verifier that combines program execution, geometric checks, and finite element analysis to provide feedback for model training.
A key learning challenge is to prioritize joint satisfaction of geometric and physical constraints while retaining informative feedback on partial progress.
A flat aggregation of heterogeneous rewards can obscure this priority, allowing improvements in individual criteria to compensate for unmet design requirements.
We therefore introduce a Hierarchical Balanced Advantage for Group Relative Policy Optimization, grouping candidates into levels according to their verification status and combining between-level progress with objective-wise peer ranking within each level.
This formulation preserves cross-level priority in the training signal while providing fine-grained feedback for satisfying constraints and improving feasible designs.
Experiments across multiple language models show that PhyCAD consistently improves joint geometric and physical constraint satisfaction.
These results demonstrate the value of verifiable engineering feedback for advancing language-driven CAD generation beyond geometric reconstruction toward physics-aware design. The code is available at \url{https://github.com/PhysAGI/PhyCAD}.

\end{abstract}

\section{Introduction}

CAD is a fundamental component of modern engineering, providing editable and parametric representations for downstream simulation, optimization, and manufacturing. Learning-based CAD generation has progressively moved from modeling CAD construction sequences and design histories~\citep{wu2021deepcad,willis2021fusion} toward language-driven generation with large language models (LLMs)~\citep{zhang2026large}. Recent methods can translate natural-language descriptions into parametric CAD sequences~\citep{khan2024text2cad}, generate CAD programs from textual or multimodal inputs~\citep{li2025llm4cad,sun2025large}, or synthesize executable modeling code through iterative or tool-augmented workflows~\citep{badagabettu2024query2cad,xie2025text,mallis2025cad}. In particular, CadQuery provides an executable, Python-based parametric representation that aligns naturally with code-generation capabilities of pretrained LLMs~\citep{xie2025text,guan2026cad}.

Despite this progress, geometric correctness alone does not ensure physical feasibility. Existing Text-to-CAD systems are commonly optimized or evaluated using geometric similarity or visual consistency~\citep{khan2024text2cad,alrashedy2025generating,xie2025text,guan2026cad}. 
However, natural-language descriptions do not always fully specify the dimensions, topology, and structural details required for engineering design, and even small geometric deviations may substantially affect structural behavior. More importantly, engineering design is ultimately concerned with whether a component meets prescribed dimensional and physical requirements under specified material properties, loads, and boundary conditions, rather than merely whether it resembles a target shape.
Bringing an initially generated model into compliance often requires repeated manual measurement, geometric modification, simulation, and re-evaluation, resulting in a costly refinement process. These limitations have motivated increasing attention to functional and physics-based criteria for evaluating and refining generated CAD models, allowing engineering feedback to guide geometry toward designs that satisfy downstream requirements~\citep{berger2026physics,son2026self}.

Computer-aided engineering (CAE), particularly simulation based on the finite element method (FEM), provides a natural source of deterministic feedback for evaluating generated designs. Recent studies have demonstrated the use of LLMs in automated FEM-based workflows and the acquisition of verifiable engineering signals~\citep{hou2025autofea,qi2025feagpt}. Simulation outcomes have also been used to guide iterative CAD generation and optimization~\citep{berger2026physics,son2026self,deng2026cosmo}. Reinforcement learning (RL) has further enabled CAD-generation models to learn from geometric feedback~\citep{guan2026cad}. Training with both geometric and physical feedback introduces a key challenge: prioritizing joint constraint satisfaction while retaining informative feedback on partial progress. A flat aggregation of heterogeneous rewards can obscure this priority, allowing improvements in individual criteria to compensate for unmet design requirements. This calls for a learning signal that distinguishes progress toward feasibility from improvements within the feasible design space.

As illustrated in Fig.~\ref{fig:teaser}, we present PhyCAD, a physics-aware Text-to-CAD framework that generates executable CadQuery programs from geometric specifications and physical constraints under prescribed material properties, loads, and boundary conditions. The task defines a constrained design space with admissible dimensional ranges rather than a unique target geometry. We construct a corresponding dataset and an automated CAD--CAE verifier that provides training feedback through program execution, geometric checks, and FEM-based simulation. Building on supervised fine-tuning, we introduce a Hierarchical Balanced Advantage for Group Relative Policy Optimization (GRPO)~\citep{shao2024deepseekmath}, grouping candidates by verification status and combining between-level progress with objective-wise peer ranking within each level. This formulation preserves cross-level priority in the training signal while providing fine-grained feedback for satisfying constraints and improving feasible designs. The trained model directly generates programs without an iterative simulation-guided refinement loop at inference time. Experiments across multiple language models show that PhyCAD consistently improves joint geometric and physical constraint satisfaction.

Our contributions are summarized as follows:
\begin{itemize}
[leftmargin=*, itemsep=0.15em, topsep=0.1em, parsep=0pt, partopsep=0pt]
\item We formulate physics-aware Text-to-CAD generation and construct a dataset that jointly captures geometric specifications, loading and boundary conditions, and physics-based constraints.
\item We develop an automated CAD--CAE verification pipeline that provides verifiable feedback on program executability, geometric and dimensional compliance, and physical constraint satisfaction to support model training.
\item We introduce a Hierarchical Balanced Advantage that combines between-level progress with objective-wise peer ranking within each level, preserving cross-level priority in the training signal while providing fine-grained feedback for satisfying constraints and improving feasible designs.
\end{itemize}

\section{Related Work}

\paragraph{Text-to-CAD Generation.}
Early learning-based CAD methods modeled construction histories, sketch constraints, and assembly relations using datasets such as DeepCAD~\citep{wu2021deepcad}, SketchGraphs~\citep{seff2020sketchgraphs}, and Fusion 360 Gallery~\citep{willis2021fusion,willis2022joinable}. Subsequent approaches map text or engineering drawings to CAD sequences using task-specific transformers~\citep{khan2024text2cad,qin2025drawing2cad}, or adapt pretrained LLMs and VLMs to structured CAD generation~\citep{li2025cad,govindarajan2025cadmium,yuan2024openecad}. CadQuery supports executable program generation~\citep{xie2025text}, while tool augmentation, self-refinement, and evolutionary search enable iterative refinement~\citep{mallis2025cad,li2026seek,preintner2025evocad}. PhyCAD studies generation under joint geometric and physical requirements, including global dimensional bounds and FEM-verified physical constraints.

\paragraph{Physics-Aware CAD Generation and Engineering Agents.}
Recently, \citep{berger2026physics} integrates physics-based verification into multi-agent CAD refinement, while \citep{son2026self} combines structured blueprints, visual inspection, and finite-element feedback to refine single-part and multi-part STEP models. COSMO-Agent~\citep{deng2026cosmo} trains an LLM through tool-augmented RL to optimize parameterized components through CAE-guided updates to geometry and material choices. Related agents automate computational fluid dynamics~\citep{chen2024metaopenfoam,xu2025cfdagent,yue2025foam}, finite-element analysis~\citep{hou2025autofea,qi2025feagpt}, and multiphysics simulation~\citep{zhang2025mooseagent}. Complementary to these engineering workflows, PhyCAD addresses a learning challenge in physics-aware CAD generation: prioritizing joint geometric and physical feasibility while retaining informative feedback on partial progress. We introduce a Hierarchical Balanced Advantage that combines progress across verification levels with objective-wise peer ranking within each level. This organizes heterogeneous CAD--CAE outcomes into a training signal that preserves cross-level priority while guiding constraint satisfaction and improvements within the feasible design space.

\paragraph{Reinforcement Learning with Verifiable Feedback.}
GRPO~\citep{shao2024deepseekmath} enables policy optimization using group-relative rewards, while InternBootcamp~\citep{li2025internbootcamp} provides diverse tasks with programmatic verifiers. In CAD, CAD-Coder~\citep{guan2026cad} combines supervised fine-tuning and GRPO using geometric and format rewards, while cadrille~\citep{kolodiazhnyi2025cadrille} uses geometric feedback for multimodal reconstruction. ReCAD~\citep{li2026recad} combines parameterized-code guidance with hierarchical primitive learning, following a curriculum over increasingly complex geometric structures. These methods primarily target geometric fidelity and output validity rather than FEM-verified physical constraints. PhyCAD incorporates CAD--CAE feedback into policy training through a Hierarchical Balanced Advantage. Rather than scheduling a geometric curriculum, it groups candidates by verification status and combines between-level progress with objective-wise peer ranking within each level.

\section{Method}
\begin{figure}[h]
    \centering
    \includegraphics[width=\textwidth]{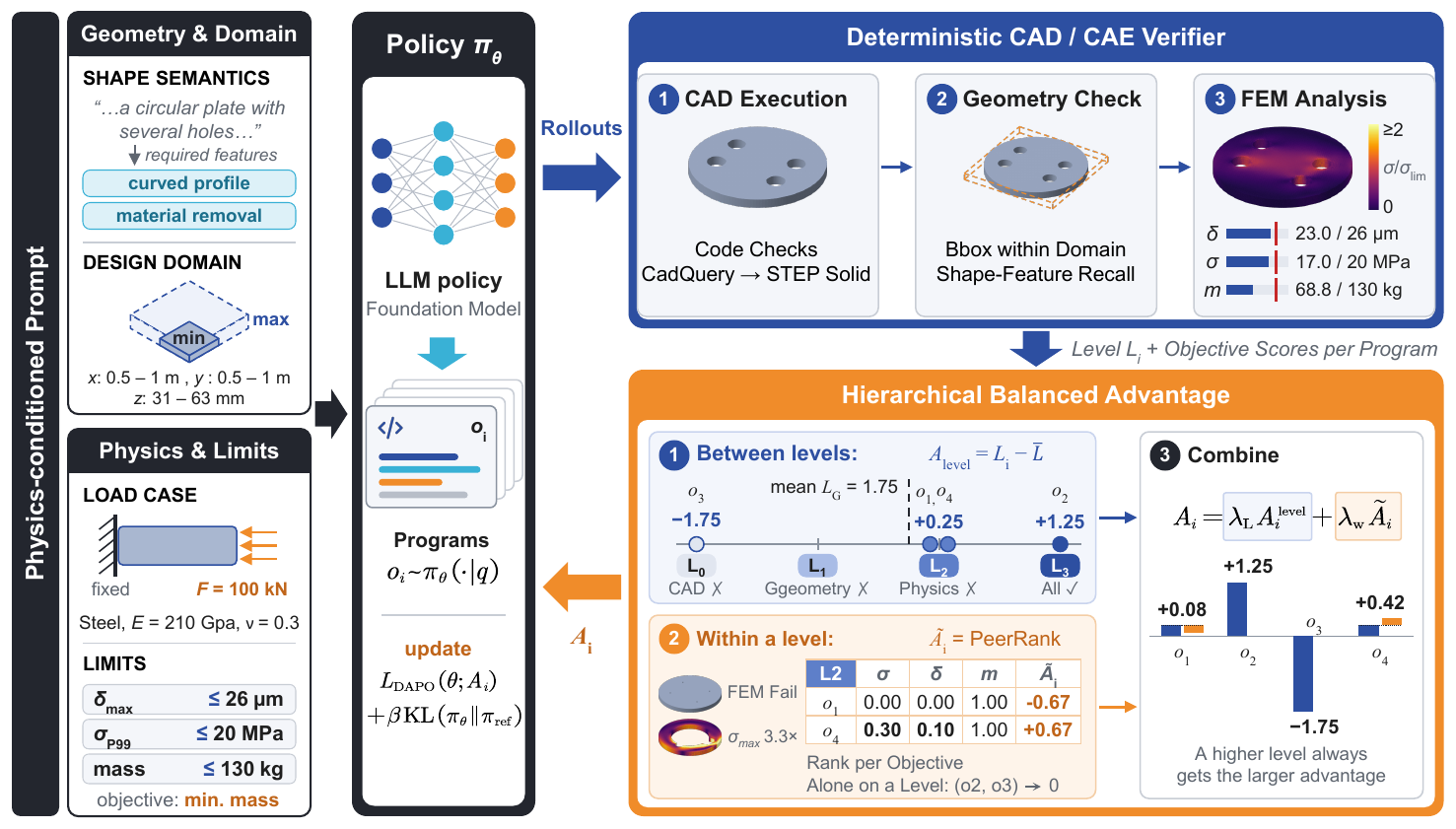}
    \caption{
    \textbf{Overview of our framework.}
    The LLM policy generates candidate CAD programs from a physics-conditioned prompt containing geometric requirements, design-domain constraints, loads, and physical limits.
    A deterministic CAD-CAE verifier sequentially evaluates each program for {CAD executability}, {geometric validity}, and {FEM-based physical feasibility and objectives}.
    These verification results define a {Hierarchical Balanced Advantage}, which combines between-level progress with within-level peer ranking over objective scores.
    The resulting advantage $A_i$ provides structured reinforcement signals for updating the CAD-generation policy.
    }
    \label{fig:overview}
    \vspace{-1em}
\end{figure}

Our goal is to train a LLM to generate CAD programs that satisfy both geometric specifications and physical constraints. As illustrated in Fig.~\ref{fig:overview}, our framework takes a physics-aware prompt as input, generates candidate CadQuery programs with an LLM policy, evaluates them through a deterministic CAD--CAE verification pipeline, and converts the verification outcomes into structured reinforcement signals for policy optimization. The resulting training framework integrates CAD executability, geometric compliance, and physics-based feasibility into a unified optimization process.

\subsection{Problem Formulation} 
\label{sec:problem_formulation}

Let \(q\in\mathcal{Q}\) denote a physics-aware design prompt describing geometric requirements, the admissible design domain, material properties, loading and boundary conditions, engineering constraints, and a design objective. An LLM policy \(\pi_{\theta}\), parameterized by \(\theta\in\mathbb{R}^{d}\), defines a conditional distribution over CadQuery programs \(\mathcal{O}\). During training, we sample a group of candidates
\(o_i\sim\pi_{\theta}(\cdot\mid q)\), \(i=1,\ldots,G\).

Our setting specifies selected geometric parameters through admissible ranges rather than fixed values, defining a constrained design space rather than a unique target geometry. A program \(o\in\mathcal{O}\) is feasible if it executes successfully to produce a valid CAD solid, satisfies the specified geometric requirements, and passes FEM evaluation with all physical constraints satisfied. We denote the feasible set by \(\mathcal{F}(q)\subseteq\mathcal{O}\). Among feasible programs, we optimize an engineering objective \(J(o;q)\), with smaller values preferred. For the tasks considered here, \(J(o;q)=m(o;q)\), where \(m(o;q)\) is the mass of the generated solid. The resulting constrained design problem is
\begin{align}
    o^{*}\in\underset{o\in\mathcal F(q)}{\arg\min}\;J(o;q).
    \label{eq:problem}
\end{align}
During training, optimization is performed over the sampled candidate set
\(\mathcal{O}_{\theta}(q)\). The CAD--CAE verifier identifies the feasible subset
\(\mathcal{F}_{\theta}(q)\) and provides hierarchical feedback for candidates
according to their verification status. The policy is optimized to increase the
probability of feasible designs while improving \(J(o;q)\) among feasible candidates.

\subsection{Deterministic CAD--CAE Verifier}
\label{sec:verifier}

For any group of programs \(o_i\) sampled from \(\pi_{\theta}(\cdot \mid q)\), we evaluate each program using the same deterministic CAD--CAE pipeline. The verifier proceeds through three stages.

\paragraph{CAD execution.}
The generated program is checked for syntax and safety, executed, and tested for valid output construction. It must produce a valid CAD solid that can be exported in STEP format and reimported by the downstream geometry kernel. Failure at this stage is recorded as a CAD-construction failure.

\paragraph{Geometry check.}
For successfully constructed solids, we compare the global bounding-box extents with the prompt's dimensional ranges and check whether the required geometric constructions are present. This checks compliance with the design requirements, separately from the solid-validity check above. Candidates satisfying these requirements proceed to CAE verification.

\paragraph{CAE verification.}
The solid is meshed and analyzed using FEM under the prescribed material properties, loads, and boundary conditions. For candidate \(o_i\), we record the volume-weighted 99th-percentile element von Mises stress \(\sigma_i\), maximum nodal displacement \(\delta_i\), and mass \(m_i=m(o_i;q)\). Let \(\bar\sigma\), \(\bar\delta\), and \(\bar m\) denote their respective upper bounds in the prompt. Physical verification succeeds when the CAE procedure completes and \(\sigma_i\leq\bar\sigma\), \(\delta_i\leq\bar\delta\), and \(m_i\leq\bar m\).

To summarize these outcomes, let \(L_i\in\{0,1,2,3\}\) denote the verification level of candidate \(o_i\):

\begin{equation}
L_i=
\begin{cases}
0, & \text{CAD construction fails},\\
1, & \text{CAD succeeds but geometric requirements fail},\\
2, & \text{geometry passes but CAE or a physical constraint fails},\\
3, & \text{all geometric and physical requirements pass}.
\end{cases}
\label{eq:level}
\end{equation}

Thus, the three checks yield four possible outcomes: failure at any one of the checks, or success at all three. A larger \(L_i\) indicates further progress through the verification procedure; \(L_i=3\) identifies a jointly feasible candidate. 

\subsection{Hierarchical Balanced Advantage} \label{sec:hba} 
The verification levels distinguish coarse progress toward feasibility but do not distinguish candidates with the same status. For example, two physically infeasible candidates may violate their constraints by very different amounts. Conversely, directly aggregating all objective scores can allow improvements in individual criteria to compensate for unmet requirements. We therefore combine a between-level advantage with a within-level comparison, retaining both the verification ordering and feedback on partial progress.

\paragraph{Between levels.} For \(G\) rollouts generated from the same prompt, we define the level advantage as 
\begin{equation} 
A_i^{\mathrm{level}} = L_i - \frac{1}{G} \sum_{j=1}^{G} L_j. 
\label{eq:level_advantage} 
\end{equation} 
This term compares each candidate's level with the group mean. Candidates above the mean receive positive level advantages. When all candidates occupy the same level, this term is zero, and the within-level component provides the remaining distinctions.

\begin{figure}[t]
    \centering
    \includegraphics[width=\textwidth]{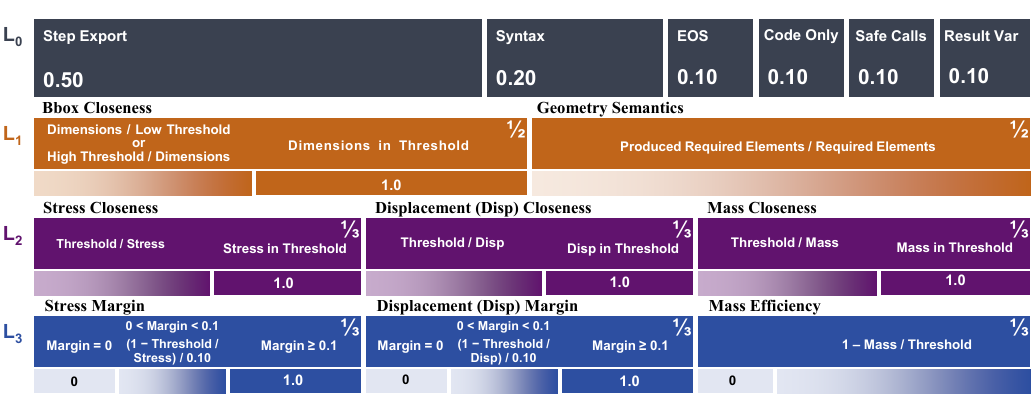}
    \caption{
\textbf{Reward definitions for within-level PeerRank.}
Each block specifies how an evaluated outcome is mapped to an objective reward
within its corresponding level.
Solid regions denote fixed rewards once a condition is satisfied, while shaded
gradients indicate rewards that vary continuously with closeness to the target
or the achieved margin; faded regions correspond to outcomes receiving zero or
lower reward.
The resulting objective rewards are independently peer-ranked and combined
using the indicated within-level weights.
}
    \label{fig:reward}
    \vspace{-1em}
\end{figure}

\paragraph{Within a level.}
Candidates are compared only with others from the same prompt and the same verification level. We first map the verification outcomes to level-specific objective scores, then rank candidates separately for each objective. Fig.~\ref{fig:reward} summarizes the score definitions and weights.
\textbf{(1).} For candidates with \(L_i=0\), CAD-progress feedback reflects output completeness, code-only formatting, syntax validity, safe calls, result construction, and STEP export. This distinguishes partial construction progress even when a valid CAD solid has not been obtained.
\textbf{(2).}  For \(L_i=1\), we assign equal weights to bounding-box closeness and geometric-semantic recall. These scores measure proximity to the dimensional ranges and coverage of the required geometric constructions, respectively.
\textbf{(3).}  For \(L_i=2\), we compare closeness to the stress, displacement, and mass limits, again with equal weights. For an available measurement \(v\) with upper bound \(\bar v\), its score is \(1\) when \(v\leq\bar v\), and \(\bar v/v\) otherwise. An unavailable measurement receives a score of zero. These scores distinguish different degrees of constraint violation without rewarding further improvement of a quantity that already satisfies its limit.
\textbf{(4).}  For \(L_i=3\), all constraints are satisfied. We compare mass efficiency \(1-m_i/\bar m\), stress margin, and displacement margin with equal weights. The two margin scores are capped at a \(10\%\) margin, beyond which additional margin receives no further reward. Feasible-design feedback therefore combines the mass objective with capped margin objectives rather than ranking feasible candidates by mass alone.

For a given prompt and a level \(\ell\in\{0,1,2,3\}\), let \(\mathcal I_\ell=\{j\in\{1,\ldots,G\}:L_j=\ell\}\) be the set of candidate indices at that level. Let \(K_\ell\) be its number of objectives. For candidate \(i\in\mathcal I_\ell\), denote its score for objective \(k\in\{1,\ldots,K_\ell\}\) by \(s_{ik}\), and its tie-aware centered rank among \(\mathcal I_\ell\) by \(r_{ik}\in[-1,1]\), with higher ranks indicating better scores. Scores within \(10^{-3}\) are treated as ties. With nonnegative objective weights \(w_{\ell k}\) satisfying \(\sum_{k=1}^{K_\ell}w_{\ell k}=1\), the within-level advantage is

\begin{equation}
\widetilde A_i=
\begin{cases}
\displaystyle\sum_{k=1}^{K_{L_i}}w_{L_i k}\,r_{ik},
& |\mathcal I_{L_i}|>1,\\[2mm]
0, & |\mathcal I_{L_i}|=1.
\end{cases}
\label{eq:peer_rank}
\end{equation}

We refer to this objective-wise ranking and averaging operation as PeerRank. It compares ranks rather than a weighted sum of the original objective scores, and gives \(|\widetilde A_i|\leq1\).


\paragraph{Combined advantage.}

The nonnegative weights \(\lambda_L\) and \(\lambda_w\) control the between-level and within-level terms, respectively. We combine them as
\begin{equation}
A_i=\lambda_L A_i^{\mathrm{level}}+\lambda_w\widetilde A_i.
\label{eq:hba}
\end{equation}

For candidates \(i\) and \(j\) from the same prompt group with \(L_i>L_j\), the common level baseline cancels, giving
\[
A_i-A_j
=\lambda_L(L_i-L_j)+\lambda_w(\widetilde A_i-\widetilde A_j)
\geq\lambda_L-2\lambda_w.
\]
Choosing \(\lambda_L>2\lambda_w\) therefore ensures that the higher-level candidate receives a larger advantage regardless of its within-level rank. This is an ordering guarantee for the training signal within each prompt group. Both components are computed together, not introduced in successive training stages.

\paragraph{Policy optimization.}
\label{sec:policy_optimization}

We use the hierarchical advantages in GRPO~\citep{shao2024deepseekmath}. The policy loss uses DAPO-style token-level aggregation: generated-token losses are averaged across the batch rather than first averaged separately within each completion. Let \(\mathbf A\) collect the advantages for the sampled completions, each computed within its own prompt group, and let \(\mathcal L_{\mathrm{DAPO}}(\theta;\mathbf A)\) denote the GRPO policy term with this aggregation. With a reference policy \(\pi_{\mathrm{ref}}\) and KL regularization weight \(\beta\geq0\), the training objective is

\begin{equation}
\mathcal L
=\mathcal L_{\mathrm{DAPO}}(\theta;\mathbf A)
+\beta D_{\mathrm{KL}}(\pi_\theta\,\|\,\pi_{\mathrm{ref}}),
\label{eq:grpo}
\end{equation}

where \(D_{\mathrm{KL}}\) denotes the Kullback--Leibler divergence. The CAD kernel and FEM solver supply verification outcomes, not gradients through geometry construction or simulation. At inference time, the trained policy directly generates a CadQuery program without an iterative simulation-guided refinement loop.

\section{Experiment}
\begin{figure}[t]
    \centering
    \includegraphics[width=\textwidth]{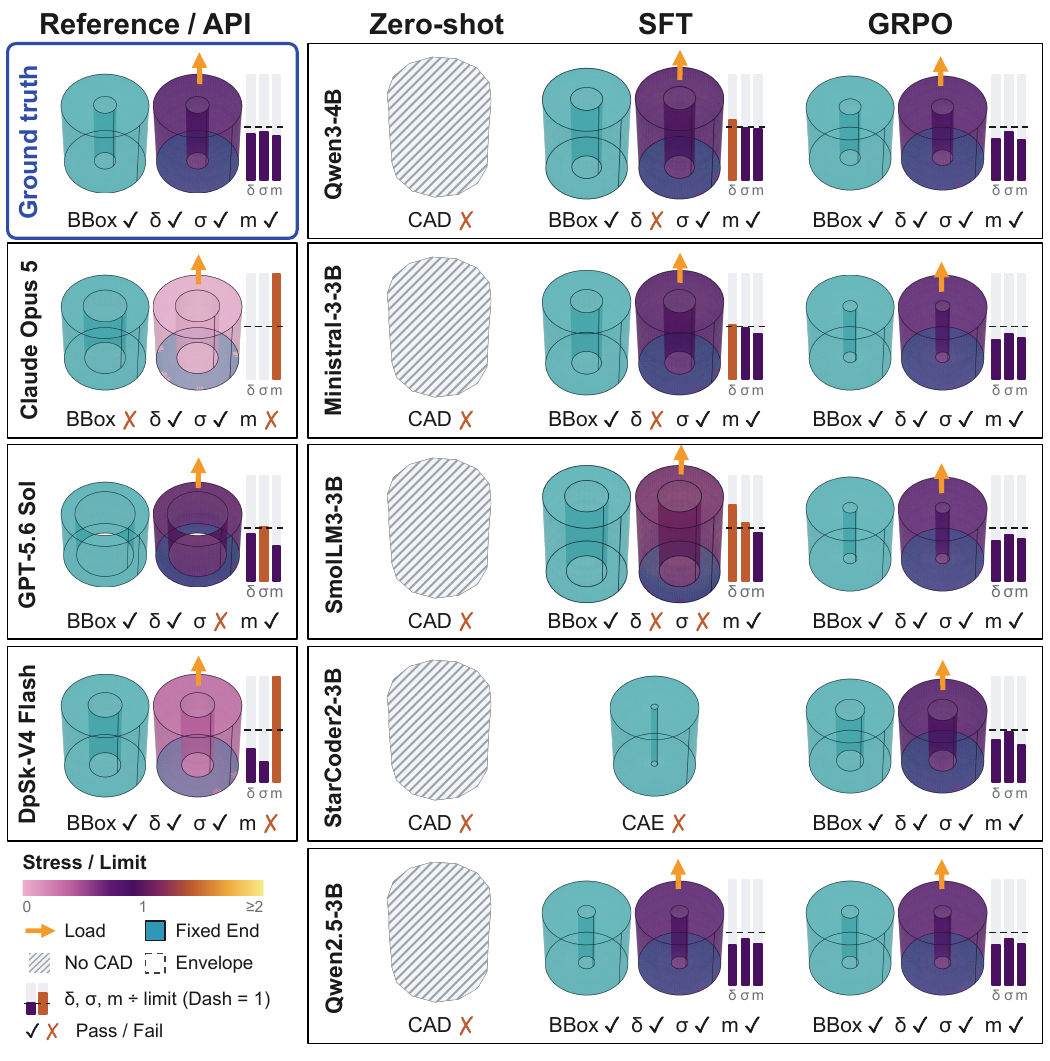}
\caption{
\textbf{Qualitative comparison across training stages and model families.}
Surface color indicates stress relative to the prescribed limit, while the bars
summarize displacement, stress, and mass constraint satisfaction.
Across model families, SFT improves CAD generation but can still violate geometric
or physical constraints, whereas GRPO more consistently achieves joint feasibility.
}
    \label{fig:qualitative_comparison}
    \vspace{-1em}
\end{figure}

\subsection{Dataset Generation}
\label{sec:dataset}

We construct a physics-aware Text-to-CAD dataset from the
Text-to-CadQuery corpus~\citep{xie2025text}, which contains approximately
\(110\text{k}\) CadQuery programs with geometric and semantic descriptions.
Each program serves as a reference design and is processed by a CAD--CAE
pipeline. We first derive boundary and loading conditions from the geometry,
then perform FEM to obtain the maximum displacement \(\delta_{\max}\),
\(99\)-th percentile von Mises stress \(\sigma_{\mathrm{P99}}\), and mass \(m\).

Each successfully analyzed reference is converted into an inverse-design task
rather than used as a direct reconstruction target. We remove exact dimensions
and other numerical specifications that could reveal the reference geometry,
while retaining high-level geometric semantics needed to define the intended
structure. The reference measurements are then used to derive displacement,
stress, and mass limits with a small feasibility margin, ensuring that the
reference remains a feasible solution without exposing its exact simulation
results. The resulting prompt specifies the geometry semantics, coarse design
bounds, boundary and loading conditions, and physical requirements, and asks
the model to generate a feasible design while minimizing mass. Detailed
construction and sanitization rules are provided in the Appendix.

After CAD--CAE annotation, \(77{,}067\) designs successfully pass the complete
pipeline. We divide them into disjoint SFT and GRPO subsets. The SFT subset contains \(11{,}024\) samples paired with sanitized
reference CadQuery programs, whereas the GRPO subset contains \(66{,}043\)
tasks without target programs and is optimized solely through CAD--CAE
verification.



\begin{table*}[t]
    \centering
    \small
    \caption{
    \textbf{Quantitative comparison on the full evaluation set of 3,373 tasks.}
    All results are reported as success rates (\%).
    Bold values indicate the best result among different settings of the same model.
    }
    \label{tab:model_comparison}
    \setlength{\tabcolsep}{5.0pt}
    \renewcommand{\arraystretch}{1.12}
    \resizebox{\textwidth}{!}{
    \begin{tabular}{llccccccc}
        \toprule
        \multirow{2}{*}{Model}
        & \multirow{2}{*}{Setting}
        & \multirow{2}{*}{Valid CAD$\uparrow$}
        & \multirow{2}{*}{BBox$\uparrow$}
        & \multirow{2}{*}{CAE$\uparrow$}
        & \multicolumn{3}{c}{Physical Constraints$\uparrow$}
        & \multirow{2}{*}{Joint$\uparrow$} \\
        \cmidrule(lr){6-8}
        & & & & 
        & Disp.
        & Stress
        & Physics
        & \\
        \midrule

        Claude Opus 5
        & Direct
        & 73.82 & 42.31 & 70.71
        & 46.52 & 49.87 & 8.42 & 7.29 \\

        GPT-5.6 Sol
        & Direct
        & 62.41 & 35.61 & 56.69
        & 46.58 & 35.75 & 16.87 & 15.51 \\

        DeepSeek-V4-Flash-Ascend
        & Direct
        & 63.33 & 50.52 & 60.45
        & 39.79 & 38.36 & 10.85 & 9.28 \\

        \midrule

        \multirow{3}{*}{Qwen3-4B}
        & Zero-shot
        & 18.56 & 11.41 & 17.55
        & 12.24 & 13.93 & 1.22 & 0.74 \\
        & SFT
        & 94.25 & 77.73 & 91.49
        & 63.09 & 63.27 & 39.19 & 36.64 \\
        & GRPO
        & \textbf{99.14} & \textbf{91.25} & \textbf{96.92}
        & \textbf{84.44} & \textbf{81.83}
        & \textbf{57.40} & \textbf{55.71} \\

        \midrule

        \multirow{3}{*}{Ministral-3-3B}
        & Zero-shot
        & 1.27 & 0.62 & 1.07
        & 0.68 & 0.65 & 0.24 & 0.12 \\
        & SFT
        & 95.55 & 78.33 & 92.68
        & 64.63 & 66.50 & 40.50 & 37.36 \\
        & GRPO
        & \textbf{98.99} & \textbf{94.22} & \textbf{97.45}
        & \textbf{84.32} & \textbf{81.20}
        & \textbf{59.29} & \textbf{58.05} \\

        \midrule

        \multirow{3}{*}{SmolLM3-3B}
        & Zero-shot
        & 6.91 & 2.91 & 6.55
        & 5.37 & 5.90 & 0.65 & 0.18 \\
        & SFT
        & 93.92 & 74.98 & 90.31
        & 58.85 & 60.72 & 31.28 & 28.73 \\
        & GRPO
        & \textbf{98.99} & \textbf{91.31} & \textbf{96.80}
        & \textbf{77.47} & \textbf{76.28}
        & \textbf{45.75} & \textbf{44.56} \\

        \midrule

        \multirow{3}{*}{StarCoder2-3B}
        & Zero-shot
        & 10.94 & 3.91 & 10.29
        & 8.81 & 8.98 & 1.04 & 0.56 \\
        & SFT
        & 93.74 & 74.50 & 90.16
        & 56.95 & 58.43 & 28.94 & 26.33 \\
        & GRPO
        & \textbf{98.96} & \textbf{92.05} & \textbf{97.15}
        & \textbf{76.67} & \textbf{74.12}
        & \textbf{43.14} & \textbf{41.65} \\

        \midrule

        \multirow{3}{*}{Qwen2.5-3B}
        & Zero-shot
        & 1.45
        & 0.47
        & 1.36
        & 1.01
        & 0.95
        & 0.21
        & 0.00 \\
        & SFT
        & 90.07 & 65.16 & 86.45
        & 55.68 & 57.75 & 25.32 & 21.76 \\
        & GRPO
        & \textbf{98.55} & \textbf{88.50} & \textbf{96.32}
        & \textbf{74.47} & \textbf{70.53}
        & \textbf{40.94} & \textbf{39.49} \\

        \bottomrule
    \end{tabular}
    }
    \vspace{-1em}
\end{table*}

\subsection{Implementation Details}
\label{sec:implementation}

From the \(66{,}043\) samples allocated to RL, we reserve \(3{,}373\) samples as a held-out evaluation set and use the remaining \(62{,}670\) prompts for GRPO training. The held-out samples are excluded from all policy updates and are used consistently for evaluating the SFT and GRPO checkpoints. All training and CAD--CAE evaluation experiments are conducted on a NVIDIA H800 GPU.

For GRPO, we use a training batch size of \(4\) prompts. For each prompt, the policy generates \(4\) independent completions, resulting in \(16\) rollout samples per batch. The four completions corresponding to the same prompt form a GRPO group and are evaluated independently by the deterministic CAD--CAE verifier described in Sec.~\ref{sec:verifier}. Their verification outcomes are then converted into the hierarchical advantages introduced above. We use a learning rate of \(5\times10^{-7}\) for policy optimization, with a sampling temperature of \(0.9\) and top-\(p\) of \(0.95\). For the hierarchical advantage, we instantiate ~\eqref{eq:hba} with
$
    \lambda_L = 1.0,\,
    \lambda_w = 0.25
$
and $\beta=0.01$ in~\eqref{eq:grpo}.


\subsection{Evaluation Metrics}
\label{sec:metrics}

We evaluate generated designs using seven metrics. \textbf{Valid CAD} measures whether the generated program successfully produces a valid CAD solid. \textbf{BBox} measures whether all three bounding-box dimensions satisfy the specified design bounds. \textbf{CAE} denotes successful finite-element analysis with valid physical measurements. \textbf{Displacement} (Disp.) and \textbf{Stress} report the pass rates for the prescribed maximum-displacement and stress constraints, respectively. \textbf{Physics} measures simultaneous satisfaction of the displacement, stress, and mass constraints. Finally, \textbf{Joint} measures whether both the geometric requirements and all physical constraints are satisfied.

\subsection{Comparison with the State of the Art}
\label{sec:sota}

We compare PhyCAD with three strong proprietary models under direct prompting,
including Claude Opus 5, GPT-5.6 Sol, and DeepSeek-V4-Flash-Ascend.
We further evaluate five open-source models under SFT and GRPO, with zero-shot
results reported where applicable, to examine the effectiveness of our training
pipeline across different model families.

Table~\ref{tab:model_comparison} and Fig.~\ref{fig:qualitative_comparison}
show a consistent progression across training stages.
SFT substantially improves CAD validity and CAE success, while GRPO further
improves geometric and physical constraint satisfaction across all evaluated
models. In particular, Joint success increases from
\(21.76\%\)--\(37.36\%\) after SFT to \(39.49\%\)--\(58.05\%\) after GRPO,
with corresponding gains in displacement and stress satisfaction.
The trained models also achieve substantially higher success on the complete
set of design requirements than proprietary models under direct prompting.
These results indicate that CAD--CAE verification provides effective supervision
beyond program validity and guides generation toward designs that better satisfy geometric and physical constraints.

\subsection{Ablation on hierarchical learning.}
Table~\ref{tab:training_stage_comparison} evaluates the effect of the hierarchical objective and staged optimization. Starting from the same SFT model, non-hierarchical GRPO improves CAD validity to $98.40\%$ but reduces joint feasibility from $21.76\%$ to $18.80\%$. In contrast, direct hierarchical GRPO achieves $98.55\%$ valid CAD and $39.49\%$ joint feasibility, nearly doubling the feasible designs relative to SFT. This indicates the gain comes from explicitly organizing feedback by feasibility level rather than from GRPO alone.

The staged results further show that optimizing geometric or physical objectives independently is insufficient: Geometry and Physics reach only $22.86\%$ and $27.63\%$ joint feasibility, respectively, improving over SFT by only $1.10$ and $5.87$ percentage points. Joint feasibility rises to $42.01\%$ only when the explicit Hierarchical objective is introduced. This is particularly important for physical feasibility, where satisfying the prescribed stress and displacement thresholds requires the model to prioritize constraint satisfaction before objective refinement. Although staged optimization yields a further $2.52$-point gain over direct hierarchical GRPO, it requires additional RL stages and substantially higher training cost.


\begin{table}[t]
\centering
\caption{
\textbf{Ablation of hierarchical reinforcement learning.} Comparison of direct hierarchical GRPO, non-hierarchical GRPO, and staged training objectives.
}
\label{tab:training_stage_comparison}
\resizebox{\linewidth}{!}{
\begin{tabular}{llcccccc}
\toprule
\textbf{Training Strategy}
& \textbf{Stage}
& \textbf{Valid CAD} $\uparrow$
& \textbf{BBox} $\uparrow$
& \textbf{CAE} $\uparrow$
& \textbf{Physics} $\uparrow$
& \textbf{Joint} $\uparrow$
& \textbf{Code Tokens} $\downarrow$ \\
\midrule

\multirow{2}{*}{Baseline}
& Zero-shot
& 1.45 & 0.47 & 1.36 & 0.21 & 0.00 & 1002.41 \\

& SFT
& 90.07 & 65.16 & 86.45 & 25.32 & 21.76 & 201.06 \\

\midrule

Direct
& Non-hierarchical GRPO
& 98.40 & 60.90 & 83.60 & 23.40 & 18.80 & 199.78 \\

\midrule

Direct
& Hierarchical GRPO
& 98.55 & 88.50 & 96.32 & 40.94 & 39.49 & 195.46 \\

\midrule

\multirow{3}{*}{Training Objective}
& $\rightarrow$ Geometry GRPO
& 95.29 & 72.37 & 91.55 & 25.88 & 22.86 & 191.60 \\

& $\rightarrow$ Physics GRPO
& 94.72 & 75.27 & 91.40 & 30.33 & 27.63 & 188.03 \\

& $\rightarrow$ Hierarchical GRPO
& \textbf{99.23}
& \textbf{92.05}
& \textbf{97.45}
& \textbf{43.34}
& \textbf{42.01}
& \textbf{184.85} \\

\bottomrule
\end{tabular}
}
\vspace{-0.8em}
\end{table}











\section{Discussion}
\paragraph{Conclusion.} Our work establishes a physics-aware Text-to-CAD setting that connects executable CAD generation with deterministic CAE verification. By combining a physics-aware dataset, a CAD--CAE verifier, and hierarchical reinforcement learning, the framework improves geometric and physical compliance across multiple models. These results demonstrate the potential of incorporating verifiable engineering feedback into CAD generation beyond geometric reconstruction.

\paragraph{Limitation and Future Work.} The current study has a few scope limitations. First, the prompts mainly encode physical constraints, while broader engineering intent is only weakly represented. Second, the dataset does not systematically cover material variations, though the same verification pipeline can be extended to different material settings. Future work could further introduce local hard dimensional constraints to enforce critical engineering requirements and reduce ambiguity in the generated geometry.

\bibliography{iclr2027_conference}
\bibliographystyle{iclr2027_conference}

\clearpage
\appendix

\section{Implementation of Dataset Generation}
\label{app:dataset}

This appendix provides implementation details for the dataset-generation
procedure introduced in Sec.~\ref{sec:dataset}. We focus on the construction
of physical conditions, inverse-design constraints, and dimension-free
geometry semantics. 

\subsection{Reference CAD--CAE Annotation and Physical Conditions}
\label{app:dataset_annotation}

Each source CadQuery program is executed and retained only if it produces
exactly one valid connected solid that can be processed by the downstream
CAD--CAE pipeline. We analyze all retained designs using three-dimensional
small-strain linear elasticity in SI units, with isotropic structural steel
($E=2.1\times10^{11}$ Pa, $\nu=0.3$, and
$\rho=7850~\mathrm{kg/m^3}$). For each successfully simulated design, we
record the maximum nodal displacement $\delta_{\max}$, the volume-weighted
99-th percentile element von Mises stress $\sigma_{\mathrm{P99}}$, and the
total mass $m$.

Because the source geometries vary substantially in shape and orientation,
the fixture and loading regions are determined automatically. We compute the
centroid of every CAD surface and select the pair of surface centroids with
the largest Euclidean separation. The normalized direction between this pair
defines the functional analysis axis. All surface centroids are projected
onto this axis, and surfaces whose projections lie within $5\%$ of either end
of the projected range form the two terminal surface groups. This grouping
allows multiple coplanar or fragmented terminal surfaces to be handled
consistently.

The negative terminal group is fully fixed
($u_x=u_y=u_z=0$), while a uniform traction is applied to the positive
terminal group such that its integrated resultant force is $100$ kN along
the positive analysis axis. The resulting analysis axis, fixture definition,
and loading condition are explicitly included in the model prompt.

\subsection{Inverse-Design Task Construction}
\label{app:inverse_task}

Each successfully analyzed reference is converted into an inverse-design task
while withholding its exact dimensions and simulation measurements.

\paragraph{Coarse design domain.}
For each global bounding-box extent $e$, we expose only the enclosing interval
between adjacent powers of two,
\begin{equation}
    B(e)
    =
    \left[
    2^{\lfloor \log_2 e \rfloor},
    2^{\lfloor \log_2 e \rfloor+1}
    \right].
    \label{eq:bbox_domain}
\end{equation}
The model therefore receives a coarse admissible range along each axis rather
than the dimensions of the reference design.

\paragraph{Performance requirements.}
The displacement, stress, and mass limits are derived independently from
$\delta_{\max}$, $\sigma_{\mathrm{P99}}$, and $m$. Each reference measurement
is multiplied by $1.10$ and rounded upward to two significant digits to form
the corresponding upper bound. The margin keeps the reference design feasible
under small numerical variations from CAD processing, meshing, and FEM
evaluation, while rounding avoids exposing its exact simulation measurements.

\begin{figure*}[t]
    \centering
    \includegraphics[width=\textwidth]{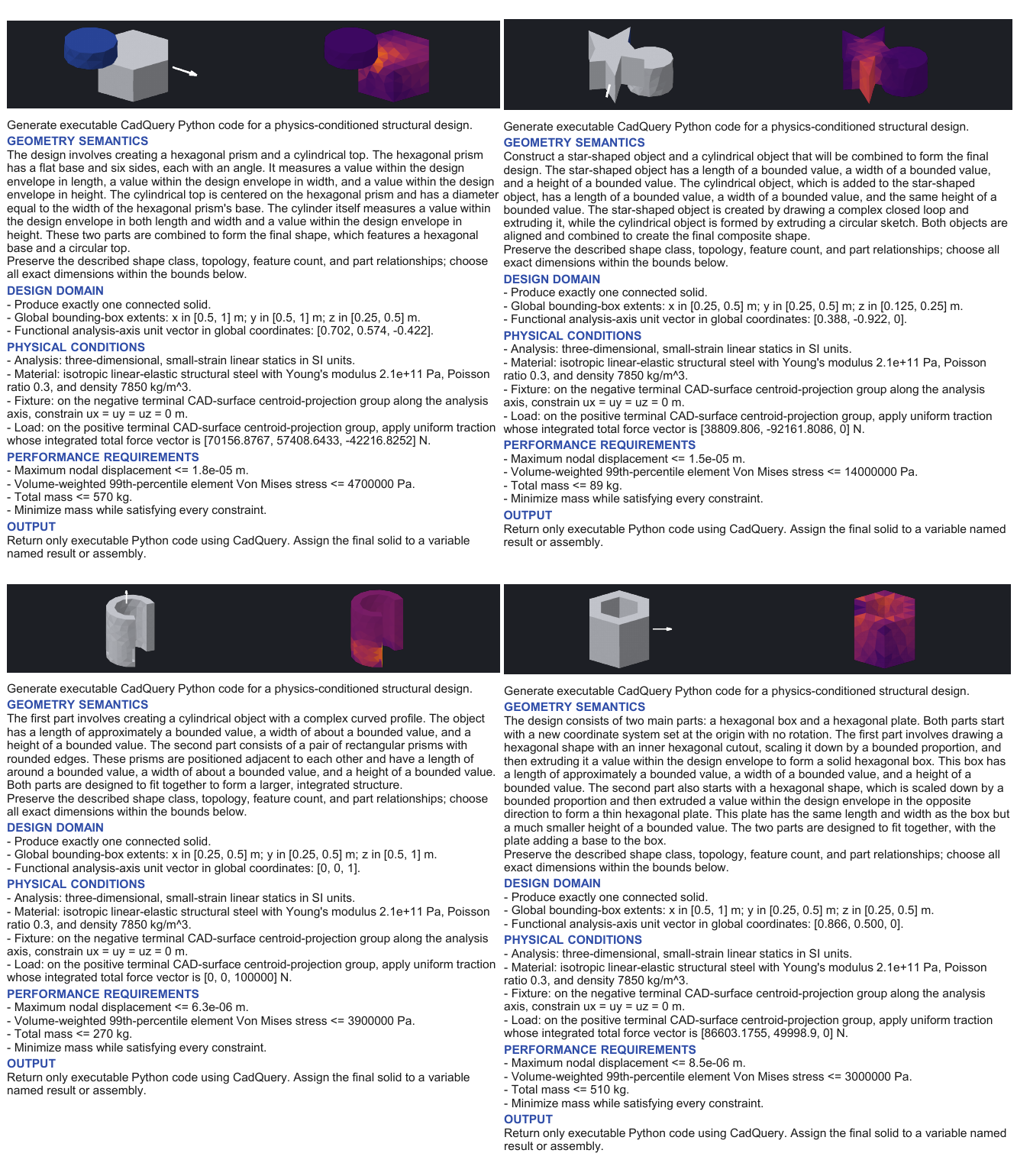}
    \caption{
    \textbf{Representative physics-aware Text-to-CAD tasks.}
    Each example shows the private reference geometry and its FEM stress field
    together with the corresponding model-visible task. Exact reference
    dimensions and simulation measurements are withheld; the model receives
    dimension-free geometry semantics, coarse design bounds, physical
    conditions, and displacement, stress, and mass limits. It must generate
    an executable CadQuery design satisfying these requirements while
    minimizing mass.
    }
    \label{fig:dataset_gallery}
\end{figure*}

\subsection{Geometry Semantic Construction}
\label{app:geometry_semantics}

The geometry specification must preserve the intended design structure without
revealing the parameters of the reference CAD. We construct this specification
from both the source textual description and the CadQuery operation sequence.

\paragraph{Numerical redaction.}
We remove numerical information that may reveal the reference geometry,
including explicit dimensions, angles, quantities with
units, and spelled-out numerical values. We retain non-numerical information
describing the shape class, topology, feature types and counts, and spatial
or compositional relationships. Thus, for example, the existence and relative
placement of a cylindrical feature are preserved while its diameter and height
are withheld.

\paragraph{Operation-derived semantics.}
Because textual descriptions do not always capture the CAD construction
consistently, we additionally parse the CadQuery program and map its modeling
operations to dimension-free geometric features. Numerical operation arguments
are discarded. Representative mappings are shown in
Table~\ref{tab:operation_semantics}.

\begin{table}[t]
    \centering
    \caption{
    Representative CadQuery operation-to-feature mappings used to construct
    dimension-free geometry semantics. Numerical parameters are not exposed
    to the model.
    }
    \label{tab:operation_semantics}
    \small
    \begin{tabular}{ll}
        \toprule
        \textbf{CadQuery construction} & \textbf{Extracted geometry semantics} \\
        \midrule
        box                         & rectangular-prismatic feature \\
        cylinder                    & cylindrical feature \\
        polygonal sketch + extrude  & polygonal-prismatic feature \\
        curved sketch + extrude     & curved-profile feature \\
        revolve                     & revolved feature \\
        cut                         & material-removal relation \\
        hole                        & hole or through-feature \\
        fillet                      & rounded-edge feature \\
        shell                       & hollow or shell construction \\
        union / composition         & part-composition relation \\
        \bottomrule
    \end{tabular}
\end{table}

The sanitized textual description and operation-derived features are combined
into the final geometry-semantic specification. It preserves the overall shape
class, topology, feature composition, and relationships between components,
while leaving their exact dimensions unspecified. Consequently, the reference
CAD remains one feasible realization of the task rather than a uniquely
specified target geometry.

Representative tasks produced by the complete pipeline are shown in
Fig.~\ref{fig:dataset_gallery}, where the reference CAD and FEM results are
included only for illustration and are not provided to the model.


\section{Hierarchical Balanced GRPO}
\label{app:hierarchical_grpo}

Algorithm~\ref{alg:hierarchical_grpo} summarizes the training procedure of our
Hierarchical Balanced GRPO. 

\begin{algorithm}[h]
\caption{\textbf{Hierarchical Balanced GRPO.}}
\label{alg:hierarchical_grpo}
\begin{algorithmic}[1]
\REQUIRE Policy $\pi_\theta$, reference policy $\pi_{\mathrm{ref}}$,
prompts $\mathcal{Q}$, group size $G$, weights $\lambda_L,\lambda_w,\beta$

\FOR{each training iteration}
    \STATE Sample a prompt batch $\mathcal{B}\subset\mathcal{Q}$

    \FOR{each prompt $q\in\mathcal{B}$}
        \STATE Sample $\{o_i\}_{i=1}^{G}\sim\pi_\theta(\cdot\mid q)$

        \STATE \textbf{Verification:}
        evaluate each $o_i$ with the CAD--CAE verifier to obtain
        level $L_i\in\{0,1,2,3\}$ and level-specific scores
        $\{s_{ik}\}_{k=1}^{K_{L_i}}$
        \STATE \hspace{\algorithmicindent}
        $L_i=0$: invalid CAD;\;
        $L_i=1$: geometry failure;\;
        $L_i=2$: physics failure;\;
        $L_i=3$: fully feasible

        \STATE \textbf{Between-level advantage:}
        $\bar L=\frac{1}{G}\sum_{j=1}^{G}L_j$,\quad
        $A_i^{\mathrm{level}}=L_i-\bar L$

        \STATE \textbf{Within-level PeerRank:}
        initialize $\widetilde A_i=0$ for all $i$
        \FOR{$\ell=0,1,2,3$}
            \STATE $I_\ell=\{i:L_i=\ell\}$
            \IF{$|I_\ell|>1$}
                \STATE Compute tie-aware centered ranks
                $r_{ik}$ of $\{s_{ik}\}_{i\in I_\ell}$
                for $k=1,\ldots,K_\ell$
                \STATE $\widetilde A_i
                =\sum_{k=1}^{K_\ell}w_{\ell k}r_{ik}$,
                \quad $\forall i\in I_\ell$
            \ENDIF
        \ENDFOR

        \STATE \textbf{Hierarchical advantage:}\quad
        $A_i=\lambda_L A_i^{\mathrm{level}}
        +\lambda_w\widetilde A_i$
    \ENDFOR

    \STATE Update $\theta$ using DAPO-style GRPO:
    $\mathcal{L}
    =\mathcal{L}_{\mathrm{DAPO}}(\theta;A)
    +\beta D_{\mathrm{KL}}(\pi_\theta\|\pi_{\mathrm{ref}})$
\ENDFOR

\RETURN Optimized policy $\pi_\theta$
\end{algorithmic}
\end{algorithm}

\end{document}